\documentclass[10pt,twocolumn,letterpaper]{article}

\usepackage{cvpr}
\usepackage{times}
\usepackage{epsfig}
\usepackage{graphicx}
\usepackage{amsmath}
\usepackage{amssymb}

\usepackage[breaklinks=true,bookmarks=false]{hyperref}

\usepackage{graphicx,float,subcaption}

\cvprfinalcopy 

\def\cvprPaperID{****} 

\begin{document}

\title{ePose: Let's Make EfficientPose More Generally Applicable}

\author{
Austin Lally\\
Oregon State University\\
{\tt\small lallya@oregonstate.edu}
\and
Robert Bain\\
Oregon State University\\
{\tt\small bainro@oregonstate.edu}
\and
Mazen Alotaibi\\
Oregon State University\\
{\tt\small alotaima@oregonstate.edu}
}

\maketitle

\begin{abstract}
EfficientPose is an impressive 3D object detection model. It has been demonstrated to be quick, scalable, and accurate, especially when considering that it uses only RGB inputs. In this paper we try to improve on EfficientPose by giving it the ability to infer an object's size, and by simplifying both the data collection and loss calculations. We evaluated ePose using the Linemod dataset and a new subset of it called "Occlusion 1-class". We also outline our current progress and thoughts about using ePose with the NuScenes and the 2017 KITTI 3D Object Detection datasets. 
The source code is available at \textcolor{blue}{\url{https://github.com/tbd-clip/EfficientPose}}.
\end{abstract}

\section{Introduction}
\label{sec:intro}
Object detection is one of the core problems in computer vision. Given an input image and a predefined object class, the task is to locate an instance of the desired object type in the image. Typically this is accomplished by reporting a bounding box. There are several variants of the problem. In multi-object detection, all instances of the object type are located. In semantic segmentation, each pixel of the image that belongs to an object of the class is labeled. Instance segmentation further distinguishes between each separate occurrence. Multi-class object detection, as the name suggests, requires locating instances of each of a set of predefined object classes instead of just one.

A natural extension of the object detection problem is to define three-dimensional bounding cubes in world coordinates instead of just drawing two-dimensional bounding boxes on the image plane. This problem, called 3D object detection, is especially important in autonomous driving and other robotics applications as agents need to predict the future positions of obstacles in 3D space. It has received a lot of attention in recent research. LiDAR-based point cloud-based methods \cite{ppc, barrera2020birdnet} have driven considerable progress. However, LiDAR sensors are not always available due to their high cost, so there is also some research into 3D object detection from RGB images \cite{wang2020pseudolidar, epose, tekin2018realtime}.

In section \ref{sec:related} we discuss related work in 3D object detection. Section \ref{sec:technical} covers the primary evaluation metric and our high-level technical approach. We describe our experiments in section \ref{sec:experiments}, and finally we summarize our thoughts and opportunities for future work in section \ref{sec:discussion}.

\section{Related Work}
\label{sec:related}
We now review existing work on 3D Object Detection that uses RGB images as an input. We divide existing detectors into two categories. We first cover traditional detectors, which require no prior information about the target object's shape, and then discuss detectors that leverage detailed 3D-models of the target object (note: we use the term ``3d-models" to distinguish mesh-style object models from deep learning models).

\subsection{Traditional Detectors}
Traditional detectors don't require 3D-models of the target objects to estimate 3D bounding cubes. Some models, like MonoPSR \cite{ku2019monocular}, use a mature 2D object detector to produce accurate 2D bounding boxes first, and then regress a 3D bounding cube from an initial 3D proposal from there. Similarly, \cite{mousavian20173d} predicts a 3D bounding cube from a 2D bounding box, but in this case they first predict the object's orientation using a novel hybrid discrete-continuous loss.

PerspectiveNet \cite{huang2019perspectivenet} instead defines ``perspective points" as the 2D projections of 3D keypoints. Using the geometric constraints imposed by the perspective projection, it reconstructs the 3D position of the target object using detected keypoints without prior knowledge about the object's 3D shape.

\subsection{Detectors Based on 3D-models}
These detectors require 3D-models of the target object to estimate the 3D bounding cubes. There are two main approaches: (i) viewpoint-based methods and (ii) keypoint-based methods. 

Viewpoint-based methods predict 3D bounding cubes with textured models. \cite{view1} proposed predicting several 3D bounding cubes for each object instance to estimate the bounding cubes distribution generated by symmetries and repetitive textures. Each predicted hypothesis corresponded to a single 3D translation and rotation, and estimated hypotheses collapsed onto the same valid bounding cube when the object appearance was unique.

Keypoint-based methods predict 3D bounding cubes by detecting specified keypoints and then solving Perspective-n-Point (PnP) problem to estimate the 3D bounding cubes. \cite{epose} proposed a 6D object pose estimation method that predicts 2D bounding boxes using a 2D object detector, uses network heads to estimate the missing depth information to lift 2D bounding boxes to 3D space, then uses PnP to refine the final pose.

The advantage of using detectors based on 3D-models is they are robust and more accurate compared to traditional detectors that don't use 3D-models to estimate the 3D bounding cubes. However, detectors based on 3D-models will require 3D-models to be used during training and evaluation of the detectors to calculate the loss rather than using the 3D bounding cube. Since ground-truth 3D-models are not always available (e.g., pedestrian detection), this is a rather limiting requirement. The other drawback to 3D-model-based detection is that it requires the detector to predict a full 3D-model, which is computationally expensive compared to simple bounding cube estimation.

\section{Technical Approach}
\label{sec:technical}

\subsection{Evaluation Metric}
\label{sec:metrics}
We evaluate our models with the ADD(-S) metric \cite{ADDS}. Symmetric and asymmetric objects are evaluated differently (ADD and ADD-S respectively). This is to ensure that the symmetric object labels are not unduly punished by a higher loss for predictions that might be semantically the same but just rotated relative to the ground truth label. This is the effect of the minimization in the for loop, which allows the comparison of object points to any other object points in $\mathcal{M}$ (originally a set of surface points from a given object of interest). We adopt the same (rather arbitrary) threshold of 10\% of the object diameter to count an object as being correctly detected or not. Said another way, if the average point pair distance between the model prediction and the ground truth is larger than 1/10th of the object's diameter, then the object was detected incorrectly.  

\vspace{0.33cm}

{\small 
$\mathrm{ADD}=\frac{1}{m} \sum_{\mathbf{x} \in \mathcal{M}}\|(\mathbf{R} \mathbf{x}+\mathbf{t})-(\tilde{\mathbf{R}} \mathbf{x}+\tilde{\mathbf{t}})\|_{2}$

\vspace{0.33cm}

$\mathrm{ADD}$-$\mathrm{S}=\frac{1}{m} \sum_{\mathbf{x}_{1} \in \mathcal{M}} \min _{\mathbf{x}_{2} \in \mathcal{M}} \|(\mathbf{R} \mathbf{x}+\mathbf{t})-(\tilde{\mathbf{R}} \mathbf{x}+\tilde{\mathbf{t}}) \|_{2}$
\par}

\subsection{Training with Bounding Cubes}

While EfficientPose \cite{epose} uses a 3D-model with many surface points $\mathcal{M}$ to model the target object, we wanted to reduce the requirement for prior information. One of our main contributions is to train the model instead on a 3D bounding cube. In this way, the loss function is computed against only eight points (instead of an average of ~500 points) per object, which reduces the number of input data points by many times.

\subsection{Improving the loss}

The other limiting assumption made by \cite{epose} is that the precise size of the target object is known ahead of time. We aim to remove this requirement and instead let the network learn the object's size. We do this by adding an additional module, identical to EfficientPose's rotation subnet, to learn a 3-dimensional scale value for the object's bounding cube. We additionally add a scaling term to the loss function in order to train the scaling subnet.

\section{Experiments}
\label{sec:experiments}
Our experiments do not focus much on the speed of training. EfficientPose \cite{epose}, which our work is built upon, also defined epochs in an odd way, which makes it difficult to compare results. Their definition is a hard-coded number of training steps per epoch, thus changing the mini-batch size or including less or more data-augmentation does not change the number of training steps in an epoch. Given more time, or if we were developing a product to be used in production, we would want to make this more in line with the standard definitions. For these reasons we will loosely talk about the time it took to train these models. The models in this report were trained on an RTX 2080 and RTX 8000 GPU with 8GB and 48GB of VRAM respectively. The latter proved useful as even small batch sizes of the smaller models ($\phi$ = 0) required a tremendous amount of VRAM to train (e.g. a batch size of 12 took 11GB). We used exclusively the smaller $\phi$ = 0 models, which took roughly 3 days each to train to completion. This included transfer learning EfficientDet \cite{eDet} weights pretrained on COCO \cite{COCO}. The rotation and translation subnetworks were always initialized from untrained weights though, as was the scaling subnetwork that will be introduced in section \ref{sec:scaling}.

The authors of \cite{epose} implemented group normalization \cite{groupnorm} in order to train large $\phi$ = 3 models, which require large amounts of VRAM to perform backprop. The experiments after and including scaling Linemod cat detector are done using batch normalization instead. Various batch sizes were used throughout training, between 1 and 50. Our intuition after tuning these models is that group norm did help at lower batch sizes, but that a batch size of 16 with the small $\phi$ = 0 models works well too. This large of a batch size was enabled by the larger memory GPU (the RTX 8000).

Table \ref{tab:adds} summarizes the best scores achieved in each of our experiments.

\subsection{Simplifying the Loss}
\label{sec:loss}

\begin{figure}[h]
    \includegraphics[width=\columnwidth]{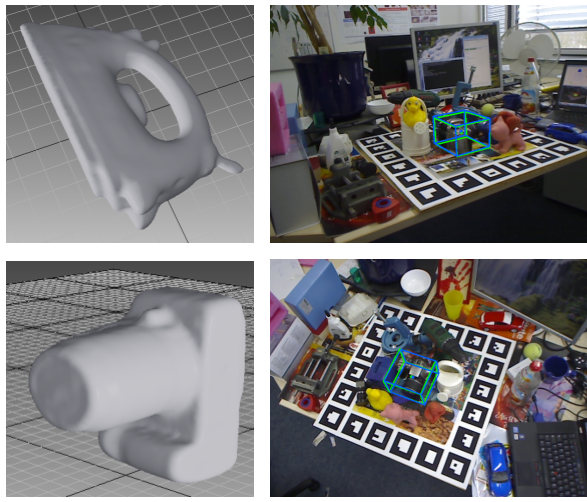}
    \caption{Results of detecting the camera (bottom left) after training a model with the bounding cube for the clothes iron (top left).}
    \label{fig:camera_iron}
\end{figure}

\begin{figure}[h]
    \includegraphics[width=\columnwidth]{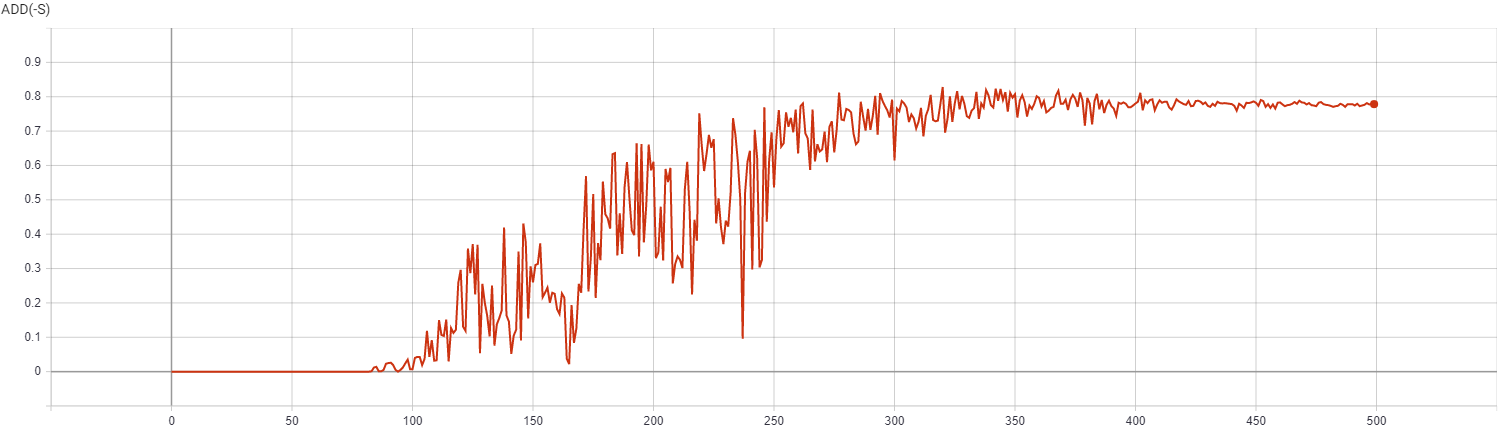}
    \caption{ADD(-S) score as a function of training epochs while training a camera detection model on a clothes iron mesh. The highest score achieved was 0.828, after 320 epochs.}
    \label{fig:training}
\end{figure}

EfficientPose \cite{epose} requires a detailed set of surface points $\mathcal{M}$ to model each potential object of interest in the image. This is a rather restricting assumption that limits the applicability to mostly toy problems. We were interested in calculating the loss without the need for this set of surface points. The intuition that motivated this experiment was that the ADD(-S) loss function would seem to be fairly agnostic to the choice of $\mathcal{M}$ (see figure \ref{fig:camera_iron} for an example). A quick test of this hypothesis was conducted by swapping the Linemod \verb`.ply` model files used to train the various single object detectors. The camera and clothes iron $\mathcal{M}$ were swapped and good results were still obtained. The ADD(-S) score was 0.83. This could have undoubtedly been improved upon, as training takes quite a while and we were convinced by the results seen in figure \ref{fig:training} that this method was working to train a detector. Thus, training was stopped before convergence so we could move on to the next experiment into simplifying the loss. 

\begin{figure*}[ht]
  \centering
  \includegraphics[]{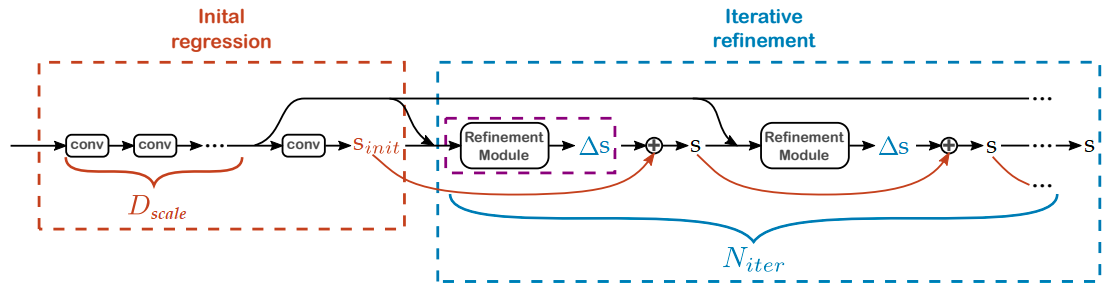}
  \caption{Our scaling subnetwork architecture. The diagram is an updated version from \cite{epose}.}
  \label{fig:scale}
\end{figure*}

We build on the previous experiment to further demonstrate that the ADD(-S) loss of \cite{epose} creates a good learning signal essentially regardless of $\mathcal{M}$. Since many 3D object data sets already have bounding cubes as part of the ground truth labels, we sought to use these in the loss function as a replacement for $\mathcal{M}$. This set of 3D points would be of length 8, reducing the number used by roughly 60x compared to \cite{epose}. This also produced good results, even though the model was trained with the 3D bounding cube vertices of the wrong \verb`.ply` model again. To reiterate: the camera detection model seen in figure \ref{fig:camera_iron} was trained with the bounding cube vertices obtained from the clothes iron \verb`.ply` model file. This model was trained until convergence and achieved 0.91 ADD(-S) (it actually did as well as 0.95 but the checkpoint directory has been lost to time). 

\begin{table}[h]
    \centering
    \begin{tabular}{|c|c|}
    \hline
        Experiment & ADD(-S) \\
    \hline
        Camera detector w/ iron model & 0.83 \\
        Camera detector w/ iron bounding cube & 0.91 \\
        Kitty detector w/scaling & 0.83 \\
        Occlusion 1-class w/bounding cube prior & 0.06 \\
        Occlusion 1-class w/o symmetric objects & 0.13 \\
    \hline
    \end{tabular}
    \caption{Best ADD(-S) achieved for each experiment.}
    \label{tab:adds}
\end{table}

\subsection{Scaling Subnetwork}
\label{sec:scaling}
Another disappointing assumption made by \cite{epose} is that oracle sizing information is known about each object of interest. This too, like the need for an accurate and detailed set of surface points, limits the applicability of this otherwise powerful method. If the scaling can instead be made part of inference, this method has a chance of working on interesting problem formulations. 

The scaling subnetwork we introduce is an exact copy of the rotation subnetwork from the original EfficientPose architecture. The architecture is shown in figure \ref{fig:scale}. The Refinement Module is just a sequence of $D_{i t e r}$ 3x3 convolutional layers. $N_{i t e r}$, $D_{i t e r}$, and $D_{r o t}$ are defined as:

\begin{equation} 
N_{i t e r}(\phi)=1+\lfloor\phi / 3\rfloor
\end{equation}

\begin{equation} 
D_{i t e r}(\phi)=D_{r o t}(\phi)=2+\lfloor\phi / 3\rfloor
\end{equation}

\begin{figure}[h]
    \centering
    \includegraphics[width=0.45\columnwidth]{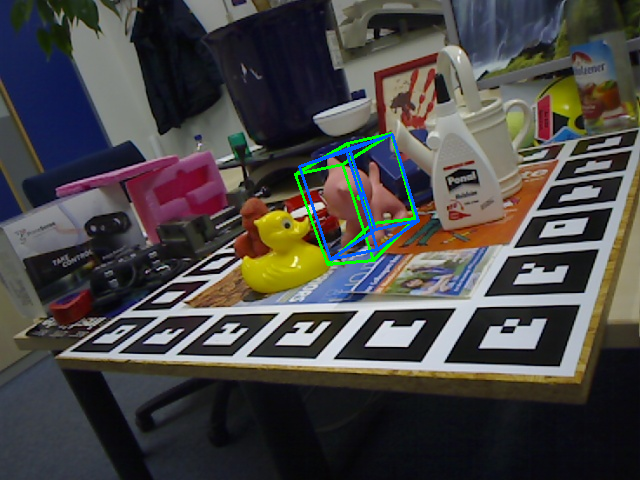}
    \includegraphics[width=0.45\columnwidth]{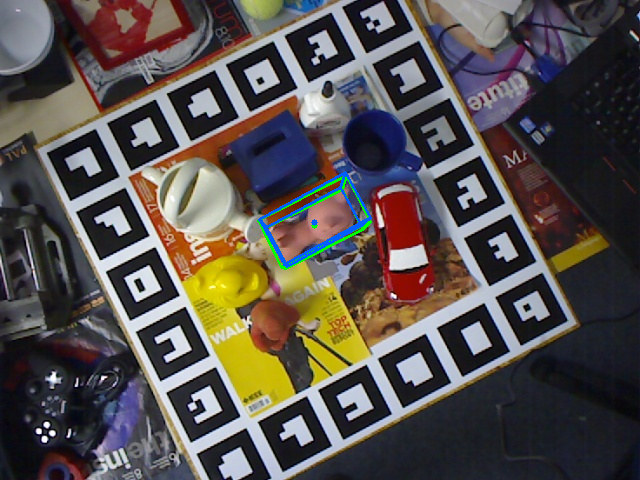}
    \includegraphics[width=0.45\columnwidth]{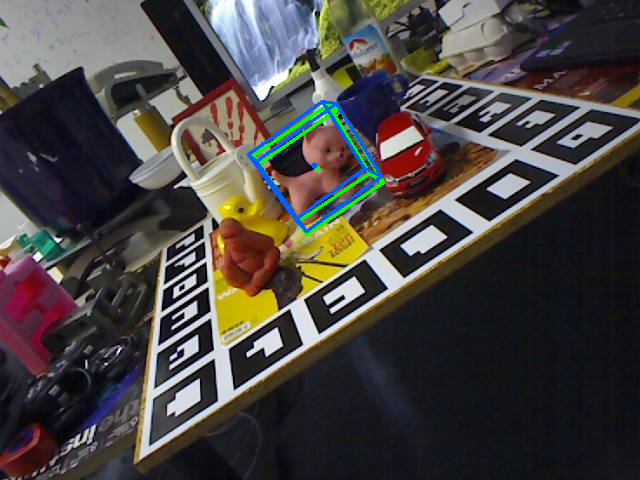}
    \includegraphics[width=0.45\columnwidth]{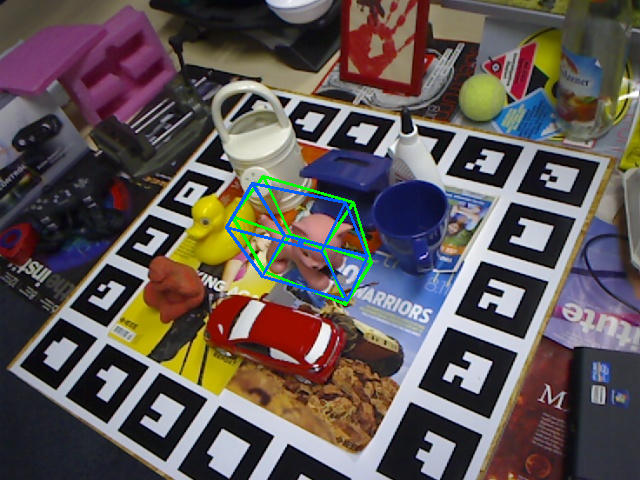}
    \caption{Sample detection results for the kitty detector with the initial proof-of-concept scaling subnet.}
    \label{fig:kitty}
\end{figure}

For this initial proof of concept the scaling subnet only had to learn to scale all the points in $\mathcal{M}$ by 1. In later experiments we use a bounding cube prior that requires more intricate scaling to be learned. The results can be seen in figure \ref{fig:kitty}. The kitty detector was also trained using the correct bounding cube vertices (not a swapped model's vertices as done in the previous subsection), and got an ADD(-S) of 0.83. It intuitively makes sense that the old performance of 0.98 ADD(-S) reported by the authors of EfficientPose should act as an upper bound on our model's performance given that they use ground truth scaling information, where instead our method must guess at that information.

Given that when \cite{epose} added the translation and rotation subnetworks to EfficientDet it only reduced inference from 35 to 27 frames per second, our addition of an exact copy of only the rotation subnetwork should have a small impact on performance.

\subsection{Occlusion 1-Class}
\label{sec:occlusion}
The occlusion dataset is a subset of the Linemod dataset. It includes a single scene with 8 annotated objects. Some of these are heavily occluded, as indicated by the name of the dataset. We made the task even more difficult by mapping all of the objects to a single class, where originally each object had its own class. This new dataset allows us to benchmark our EfficientPose additions against use cases where there are multiple objects of interest in a given image, and the use cases where there is intra-class size variance (i.e. when objects in the same class have different scalings). 

Unlike the scaling in the previous section, here we use a bounding cube prior that is an average bounding cube over all 8 objects in the class. The prior is then directly scaled using the output of the scaling subnetwork.

\begin{figure}[h]
  \centering
  \includegraphics[width=\columnwidth]{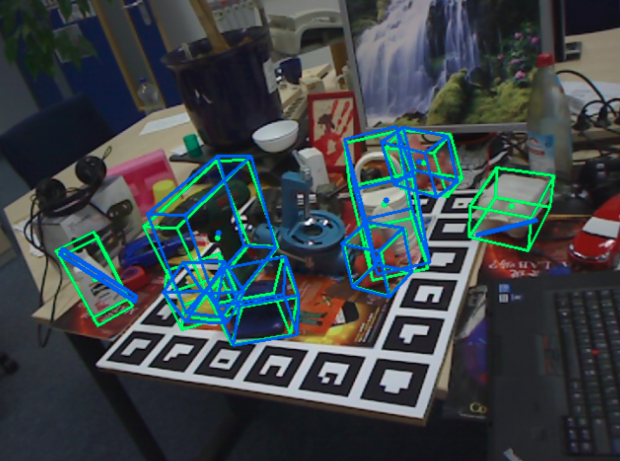}
  \caption{Example inference from our new scaling subnet addition. Note that the two worse object predictions are both symmetric objects. This highlights how adding scaling makes the ADD-S metric inappropriate for training.}
  \label{fig:occ1-symm}
\end{figure}

\begin{figure}[h]
  \centering
  \includegraphics[width=\columnwidth]{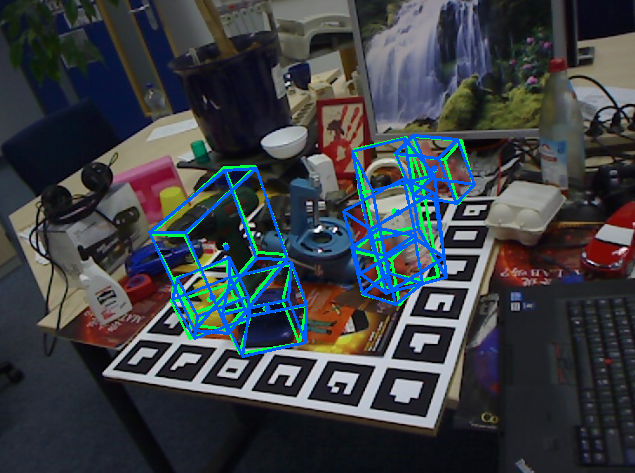}
  \caption{Results from the same experimental setup as \ref{fig:occ1-symm}, but with the symmetric objects removed from training. The ADD(-S) score then jumps from 0.06 to 0.13. This demonstrates that training our method on symmetric objects will not only fail, but also degrades the performance on other, non-symmetric objects.}
  \label{fig:occ1-no-symm}
\end{figure}

Using a class average object diameter of 170 mm led to an ADD(-S) score of only 0.06 for this difficult dataset. Upon inspecting the results, one quickly realizes that the symmetric objects are failing to be learned. The model always predicts scalings that result in small volumes. This effect can be seen in figure \ref{fig:occ1-symm}. The ADD-S metric and loss for symmetric objects was discovered to have become inappropriate for a network that includes a scaling inference, because the inner-minimization allows the most favorable points to be compared in order to avoid over-penalizing model predictions. The model then learns to make a very small volume guess that is close to just 1 or 2 points. 

We took the two symmetric objects out of the dataset and re-ran with the same hyper-parameters (\ref{fig:occ1-no-symm}). The ADD(-S) results more than doubled to 0.13 ADD(-S). This indicates that not only will we need to change the loss for symmetric objects, but currently including symmetric objects will hinder the performance w.r.t. the other asymmetric objects in the dataset.

\section{Discussion}
\label{sec:discussion}

\subsection{Challenges}
\subsubsection{Symmetry Problem}
We tried to solve the symmetry problem where our detector fails to estimate 3D bounding cubes for symmetrical objects. To solve this problem, we introduced an additional loss to the translation loss to estimate the volume of the object. We have tried two approaches to define the additional loss: a simple volume estimation and a rotation-robust IoU (RIoU) \cite{riou}. The simple volume estimation compares the volume of predicted 3D bounding cube and the target 3D bounding cube. Rotation-robust IoU \cite{riou} computes the Intersection over Union (IoU) for the 3D bounding cube by taking rotation into consideration, as defined in equation \ref{eq:iou}. 

\begin{equation}
\label{eq:iou}
\begin{split}
I_{R \operatorname{Io} U}&=\min (I 1, I 2) \cdot\left|\cos \left(2 \cdot\left(r_{g}-r_{p}\right)\right)\right| \\
U_{R \operatorname{Io} t}&=\max \left(I_{R I \mathrm{o} U}, l_{g}-w_{g}+l_{p}-w_{p}-I_{R I \mathrm{o} U}\right) \\
R \mathrm{IoU}&=\frac{I_{\mathrm{RIO} U}}{U_{\mathrm{RToU}}}
\end{split}
\end{equation}

In these equations (as described in \cite{riou}), $l_g$,$w_g$, and $r_g$ represent length, width, and rotation around z-axis respectively of the ground truth 3D bounding cube. $l_p$,$w_p$, and $r_p$ represent length, width, and rotation around z-axis respectively of the predicted 3D bounding cube. $I1$ represents the intersection of the predicted and ground truth bounding cubes in the coordinate space of the ground truth, and $I2$ represents the intersection of the predicted and ground truth bounding cubes in the coordinate space of the prediction.

Neither definition of the loss yielded good results when training on symmetric objects without using 3D-models of the objects compared to training on symmetric objects with 3D-models of objects on Linemod dataset. This means that detection of symmetric objects remains an unsolved problem in our approach, but we hope that with further research an appropriate loss function can be found to resolve the issue. We suggest an alternative approach in section \ref{sec:future}.

\subsubsection{NuScenes}
NuScenes \cite{nuscenes2019} offers an autonomous driving dataset that looks promising for our use case. It consists of 1000 20-second video scenes captured in Boston and Singapore. In each keyframe, 23 object classes are annotated with ground-truth 3D bounding cubes. In each keyframe, several sensor images are captured along with the ground truth annotations. There are six cameras, one LiDAR, five RADAR, one GPS, and an IMU. We ignore the other sensors and use only the RGB camera images for training and evaluation.

\begin{figure}[h]
    \centering
    \includegraphics[width=\columnwidth]{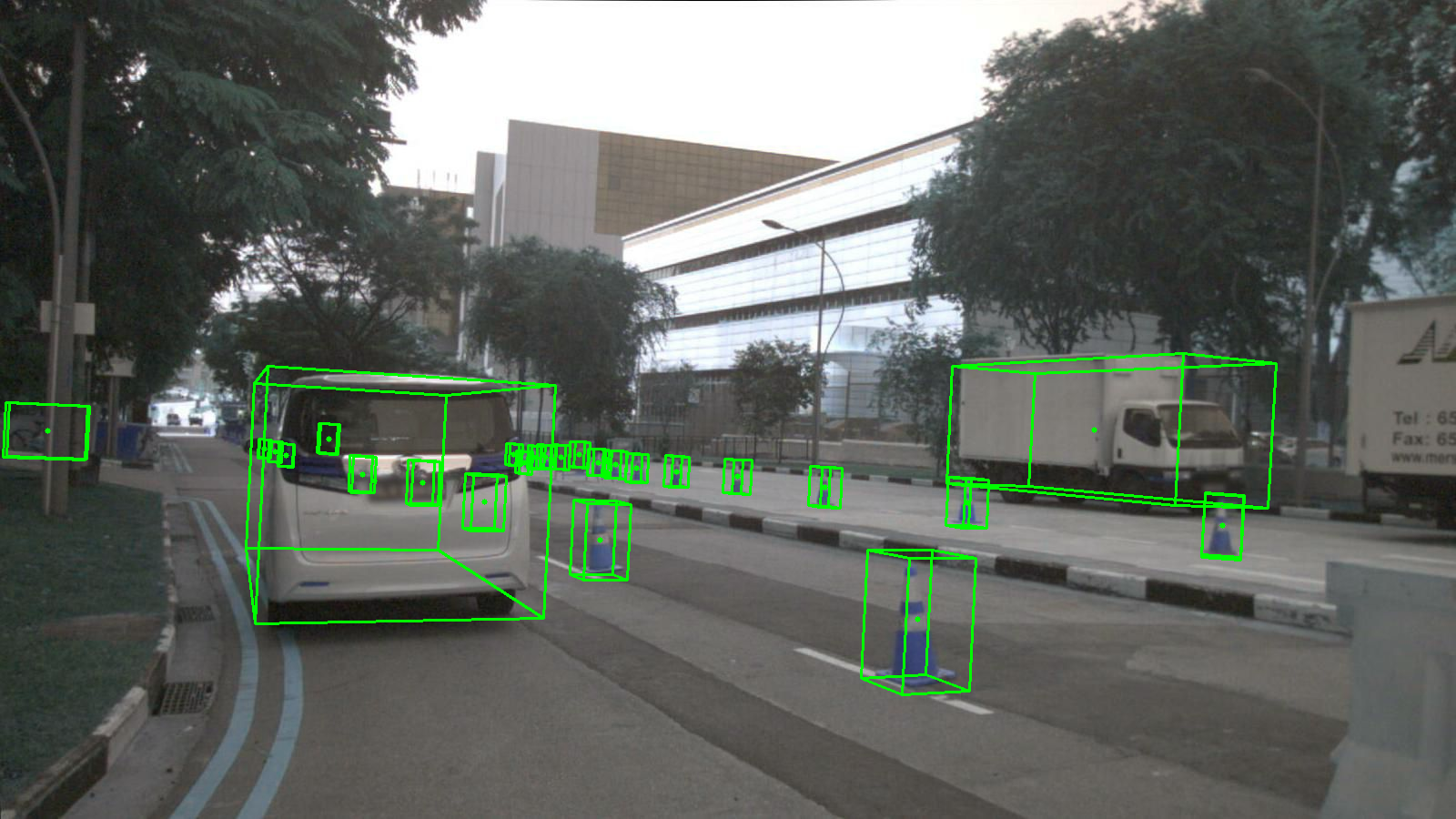}
    \includegraphics[width=\columnwidth]{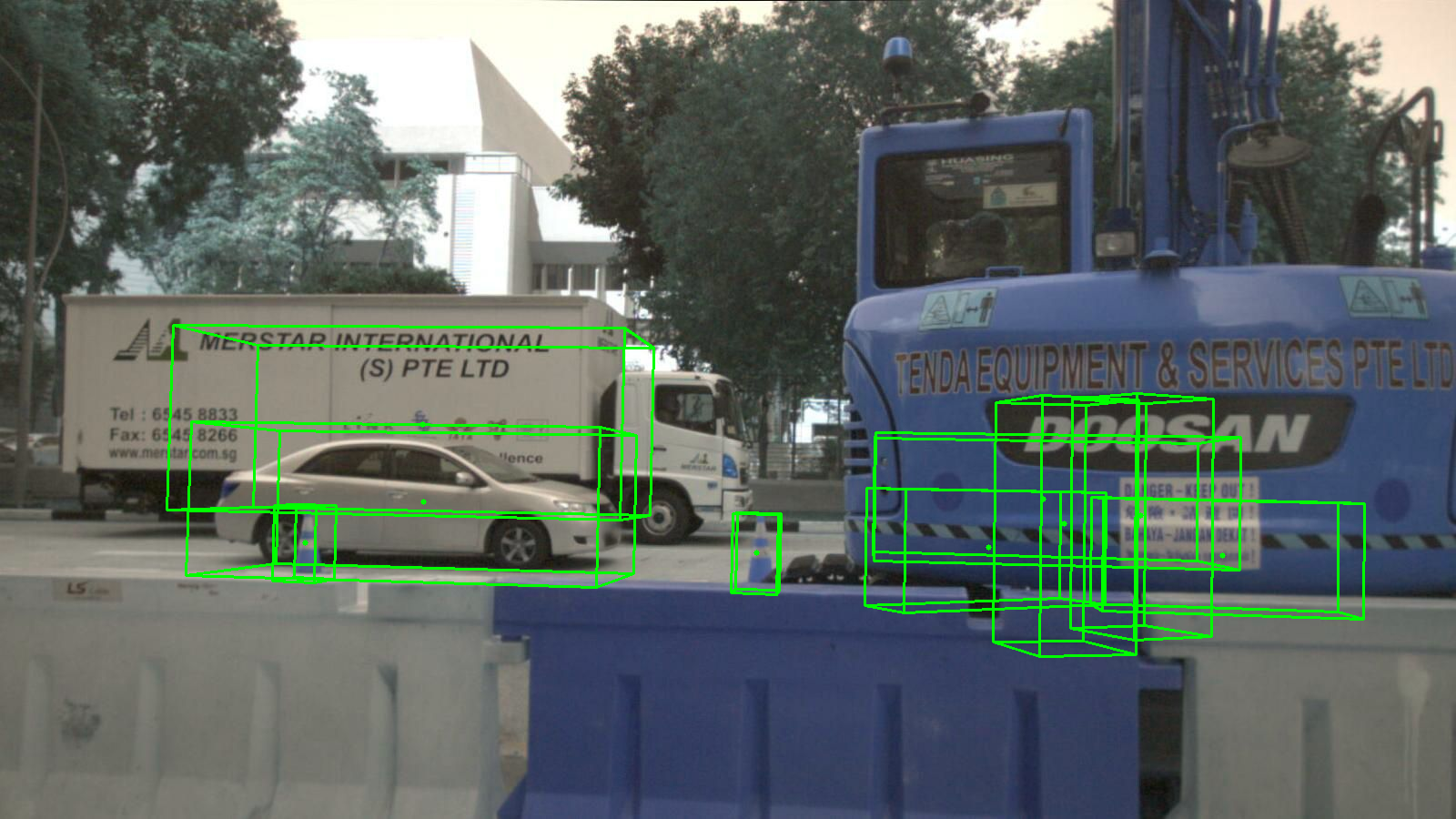}
    \caption{NuScenes images with heavily occluded ground-truth bounding cubes.}
    \label{fig:nuscenes}
\end{figure}

The ground-truth bounding cubes are given as an 8x3 matrix of coordinates, with the four front points listed first in clockwise order, followed by the back four. This fits well with the loss model described in section \ref{sec:loss}. Unfortunately, after building a data loader for this dataset and incorporating it into the EfficientPose codebase, the model was entirely unsuccessful and failed to produce almost any detections. Debugging this behavior led to the discovery that the dataset includes many fully occluded ground-truth bounding cubes, as seen in figure \ref{fig:nuscenes}. 

Two reasons are given for this in the NuScenes issue tracker. First, the LiDAR hat is mounted much higher on the vehicle than the cameras, so although objects are hidden from camera view they may be visible by LiDAR in the same keyframe. Second, NuScenes propagates \textit{temporarily} occluded bounding cubes through intermediate frames (although it's unclear under what conditions they do this, and how they define \textit{temporarily} in this case). This obviously harms training when looking only at camera data, because the model is trying to learn to detect what isn't visible.

\subsection{Future Work}
\label{sec:future}

Future researchers might be interested in a solution to the volume collapse that we came up with. Each scaling subnet output element could be put through an exponential and then offset by some small amount (e.g. 0.2). \cite{yolov3} scales its anchor bounding boxes in a similar way, but without the offset. The offset would essentially enforce a minimum volume of the predicted bounding cube, preventing the undesirable effects of the current ADD-S loss that unknowingly encourages the scaling net to predict very small volume bounding cubes.

We are still interested in applying ePose to a large dataset like NuScenes, even if NuScenes in particular is not applicable to a model like ours that does not leverage LiDAR information. We are currently in the process of evaluating ePose on the 2017 KITTI dataset for 3D Object Detection \cite{KITTI}. This dataset has multiple levels of difficulty, classified by the area of each annotation's bounding box, and only includes 1 degree of rotation about the y-axis. The 2017 KITTI dataset annotates the pose and rotation of bounding cubes centered around pedestrians, cyclists, and cars. The volume overlap threshold used for the KITTI leaderboards is 70\% for cars, and only 50\% for the other 2 classes. It would be interesting to see how our model ranks on the leaderboard, against solutions that use LiDAR or depth information.

\subsection{Conclusion}

EfficientPose \cite{epose} is a promising method for 3D object pose estimation, but it has real limitations when it comes to applying it to some real-world situations. We endeavored to improve the flexibility of the approach by reducing the amount of detailed object information necessary to train the network. We found that it is in fact possible to train the network using 3D bounding cubes instead of high-fidelity 3D-models. We also found that a scaling subnetwork successfully learns object sizes to a reasonable degree. We found that symmetric objects pose a real challenge, and that additional research is needed to get the scaling module working well in those cases. We also proposed possible future extensions to our work. We consider our work to be a substantial contribution to the field, and hope that continued work in this direction can lead to highly successful real-time 3D object detection from RGB images.

{\small
\bibliographystyle{ieee}
\bibliography{egbib}

\begin{thebibliography}{10}\itemsep=-1pt

\bibitem{barrera2020birdnet}
A.~Barrera, C.~Guindel, J.~Beltrán, and F.~García.
\newblock Birdnet+: End-to-end 3d object detection in lidar bird's eye view,
  2020.

\bibitem{epose}
Y.~Bukschat and M.~Vetter.
\newblock Efficientpose: An efficient, accurate and scalable end-to-end 6d
  multi object pose estimation approach, 2020.

\bibitem{nuscenes2019}
H.~Caesar, V.~Bankiti, A.~H. Lang, S.~Vora, V.~E. Liong, Q.~Xu, A.~Krishnan,
  Y.~Pan, G.~Baldan, and O.~Beijbom.
\newblock nuscenes: A multimodal dataset for autonomous driving.
\newblock {\em arXiv preprint arXiv:1903.11027}, 2019.

\bibitem{KITTI}
A.~Geiger, P.~Lenz, and R.~Urtasun.
\newblock Are we ready for autonomous driving? the kitti vision benchmark
  suite.
\newblock In {\em Conference on Computer Vision and Pattern Recognition
  (CVPR)}, 2012.

\bibitem{ADDS}
S.~Hinterstoisser, V.~Lepetit, S.~Ilic, S.~Holzer, G.~R. Bradski, K.~Konolige,
  and N.~Navab.
\newblock Model based training, detection and pose estimation of texture-less
  3d objects in heavily cluttered scenes.
\newblock In K.~M. Lee, Y.~Matsushita, J.~M. Rehg, and Z.~Hu, editors, {\em
  ACCV (1)}, volume 7724 of {\em Lecture Notes in Computer Science}, pages
  548--562. Springer, 2012.

\bibitem{huang2019perspectivenet}
S.~Huang, Y.~Chen, T.~Yuan, S.~Qi, Y.~Zhu, and S.-C. Zhu.
\newblock Perspectivenet: 3d object detection from a single rgb image via
  perspective points, 2019.

\bibitem{ppc}
L.~Jansen, N.~Liebrecht, S.~Soltaninejad, and A.~Basu.
\newblock 3d object classification using 2d perspectives of point clouds.
\newblock In T.~McDaniel, S.~Berretti, I.~D.~D. Curcio, and A.~Basu, editors,
  {\em Smart Multimedia}, pages 453--462, Cham, 2020. Springer International
  Publishing.

\bibitem{ku2019monocular}
J.~Ku, A.~D. Pon, and S.~L. Waslander.
\newblock Monocular 3d object detection leveraging accurate proposals and shape
  reconstruction, 2019.

\bibitem{COCO}
T.~Lin, M.~Maire, S.~J. Belongie, L.~D. Bourdev, R.~B. Girshick, J.~Hays,
  P.~Perona, D.~Ramanan, P.~Doll{\'{a}}r, and C.~L. Zitnick.
\newblock Microsoft {COCO:} common objects in context.
\newblock {\em CoRR}, abs/1405.0312, 2014.

\bibitem{view1}
F.~Manhardt, D.~M. Arroyo, C.~Rupprecht, B.~Busam, T.~Birdal, N.~Navab, and
  F.~Tombari.
\newblock Explaining the ambiguity of object detection and 6d pose from visual
  data, 2019.

\bibitem{mousavian20173d}
A.~Mousavian, D.~Anguelov, J.~Flynn, and J.~Kosecka.
\newblock 3d bounding box estimation using deep learning and geometry, 2017.

\bibitem{yolov3}
J.~Redmon and A.~Farhadi.
\newblock Yolov3: An incremental improvement.
\newblock {\em CoRR}, abs/1804.02767, 2018.

\bibitem{eDet}
M.~Tan, R.~Pang, and Q.~V. Le.
\newblock Efficientdet: Scalable and efficient object detection.
\newblock {\em CoRR}, abs/1911.09070, 2019.

\bibitem{tekin2018realtime}
B.~Tekin, S.~N. Sinha, and P.~Fua.
\newblock Real-time seamless single shot 6d object pose prediction, 2018.

\bibitem{wang2020pseudolidar}
Y.~Wang, W.-L. Chao, D.~Garg, B.~Hariharan, M.~Campbell, and K.~Q. Weinberger.
\newblock Pseudo-lidar from visual depth estimation: Bridging the gap in 3d
  object detection for autonomous driving, 2020.

\bibitem{groupnorm}
Y.~Wu and K.~He.
\newblock Group normalization.
\newblock {\em CoRR}, abs/1803.08494, 2018.

\bibitem{riou}
Y.~Zheng, D.~Zhang, S.~Xie, J.~Lu, and J.~Zhou.
\newblock Rotation-robust intersection over union for 3d object detection.
\newblock In A.~Vedaldi, H.~Bischof, T.~Brox, and J.-M. Frahm, editors, {\em
  Computer Vision -- ECCV 2020}, pages 464--480, Cham, 2020. Springer
  International Publishing.

\end{thebibliography}
}

\end{document}